\documentclass[letterpaper, conference]{ieeeconf}  
\usepackage{tipa}
\usepackage{bm}
\usepackage{url}
\usepackage{balance}
\usepackage{graphicx}
\usepackage{amsfonts}
\usepackage{fancyhdr}
\usepackage{comment}
\usepackage{times}
\usepackage{amsmath}
\usepackage{changepage}
\usepackage{amssymb}
\usepackage{enumerate}
\usepackage{algorithmic}
\usepackage{bm}
\usepackage{subfigure}
\usepackage{calligra}
\usepackage{multirow}
\usepackage{tabularx}
\usepackage{booktabs}
\usepackage{mathtools}
\usepackage{arydshln}
\usepackage{latexsym}
\usepackage{amsmath}
\usepackage{amssymb}
\usepackage{color}
\usepackage{soul}
\usepackage{gensymb}
\usepackage{cite}
\usepackage{color}
\usepackage{cuted}
\usepackage{etoolbox}
\usepackage{capt-of}
\usepackage{subfigure}
\usepackage[flushleft]{threeparttable}
\usepackage{ulem}
\usepackage{float}
\usepackage{booktabs}
\usepackage{bookmark}
\usepackage{hyperref}
\usepackage[ruled,vlined]{algorithm2e}
\makeatletter
\def\hlinewd#1{%
  \noalign{\ifnum0=`}\fi\hrule \@height #1 \futurelet
   \reserved@a\@xhline}
\makeatother
\AfterEndEnvironment{strip}{\leavevmode}

\newcommand{\redtext}{\textcolor[rgb]{1,0,0}}

\newcommand{\darkgraytext}{\textcolor[rgb]{0.4,0.4,0.4}}

\IEEEoverridecommandlockouts

\title{\LARGE \bf
Greedy-Based Feature Selection for Efficient LiDAR SLAM
\author{Jianhao Jiao$^{1}$, Yilong Zhu$^{1}$, Haoyang Ye$^{1}$, Huaiyang Huang$^{1}$, Peng Yun$^{1}$, Linxin Jiang$^{2}$, 
\\Lujia Wang$^{2}$, Ming Liu$^{1}$
}
\thanks{This work was supported by the National Natural Science Foundation of China, 
under grant No. U1713211, Collaborative Research Fund by Research Grants Council Hong Kong, 
under Project No. C4063-18G, and Zhongshan Municipal Science and Technology Bureau Fund, under project ZSST21EG06, awarded to Prof. Ming Liu.}
\thanks{$^{1}$Jianhao Jiao, Yilong Zhu, Haoyang Ye, Huaiyang Huang, Peng Yun, and Ming Liu are with the Robotics Institute, HKUST, Hong Kong SAR, China.
{\tt\small \{jjiao,yzhubr,hyeab,hhuangat,eelium\}}@ust.hk.}
\thanks{$^{2}$Lingxin Jiang and Lujia Wang are with Shenzhen Institutes of Advanced Technology, Chinese Academy of Sciences, Shenzhen, China.
{\tt\small \{lingxinjiang\}}@gmail.com, 
{\tt\small \{lj.wang1\}}@siat.ac.cn.}
}

\begin{document}
\maketitle


\begin{abstract}
Modern LiDAR-SLAM (L-SLAM) systems have shown excellent results in large-scale, real-world scenarios. 
However, they commonly have a high latency due to the expensive data association and nonlinear optimization. 
This paper demonstrates that actively selecting a subset of features significantly improves both the accuracy and efficiency of an L-SLAM  system.
We formulate the feature selection as a combinatorial optimization problem under a cardinality constraint to preserve the information matrix's spectral attributes.
The stochastic-greedy algorithm is applied to approximate the optimal results in real-time. 
To avoid ill-conditioned estimation, we also propose a general strategy to evaluate the environment's degeneracy and modify the feature number online.
The proposed feature selector is integrated into a multi-LiDAR SLAM system.
We validate this enhanced system with extensive experiments covering various scenarios on two sensor setups and computation platforms.
We show that our approach exhibits low localization error and speedup compared to the state-of-the-art L-SLAM systems. 
To benefit the community, we have released the source code: \url{https://ram-lab.com/file/site/m-loam}.
\end{abstract}

\section{Introduction}

\subsection{Motivation}
State estimation is a classic and fundamental problem in robotics \cite{cadena2016past}.
Over the past decades, LiDARs have attracted much attention from the simultaneous localization and mapping (SLAM) community due to their accuracy and reliability in range measurements.
Recent work \cite{zhang2014loam,shan2018lego,behley2018efficient,lin2020decentralized,liu2020balm} has pushed LiDAR-SLAM (L-SLAM) systems that are accurate and robust.
However, L-SLAM systems commonly present a high latency on a variety of on-board processors with limited computation resources.
This issue is critical if the scale of SLAM becomes large, or modules such as high-level decision making are integrated. 
Thus, towards real-time SLAM for diverse applications, such systems must exhibit low latency (time delay between input and output) along with the preservation of their accuracy and robustness.

L-SLAM comprises two major computational tasks in L-SLAM: 
data association and optimization.
Data association indicates feature matching between the current frames to the reference frames, while optimization solves the pose parameters by maximizing a likelihood function given a set of constraints.
Compared to visual features such as SIFT \cite{lowe2004distinctive} and ORB \cite{rublee2011orb}, matching 3D features is known to be less accurate \cite{yang2019polynomial}, thus producing much higher outlier rates. 
To enforce accuracy, 
most L-SLAM systems exploit thousands of features to solve a large nonlinear least-squares (NLS) problem. 
However, this scheme presents significant drawbacks. 
The data association has to perform numerous nearest-neighbor queries to match correspondences, which is commonly time-consuming.
Given plentiful measurements, the computational complexity of the optimization based on gradient descent also grows quadratically.

A prevalent solution to bound the complexity is to perform data sampling.
For instance, many LiDAR-based object detectors \cite{qi2017pointnet,qi2017pointnet++,shi2018pointrcnn} leverage the farthest point sampling (FPS) \cite{eldar1997farthest} to process input points.
However, classic sampling methods do not consider downstream tasks specifically.
Taking FPS as an example, it selects a subset of points with the objective to achieve a maximal coverage of the input set \cite{lang2020samplenet}.
But for SLAM, it would be better if the process of point selection conforms to the optimization objective.
Ideally, exploiting the set of selected features in optimization should lead to low latency and performance improvements. 

\subsection{Contributions} 
This paper proposes a general and straightforward feature selection algorithm for L-SLAM systems.
We have a crucial observation behind our approach: that not all geometric constraints contribute equally to the localization accuracy.
Intuitively, well-conditioned constraints should distribute different directions, 
constraining the pose from different angles \cite{zhang2016degeneracy}.
For instance, orthogonal constraints commonly outweigh their parallel counterparts.
The selected features, which are the most valuable/informative to the pose estimation, are defined as \textbf{good features} \cite{zhao2020good}, 
and both the data association and state optimization utilize them only.

This paper extends our previous work on multi-LiDAR SLAM \cite{jiao2020robust}. 
Multiple LiDARs enable a robot to maximize its perceptual awareness of environments and obtain sufficient measurements, but inevitably increase the processing time.
In this paper, we investigate the latency issue.
From the traditional perspective, there is a trade-off between the latency and accuracy \cite{bodin2018slambench2}. 
But in Section \ref{sec.experiment}, 
we demonstrate that the proposed feature selection method 
boosts the accuracy ($>22\%$ error reduction) and efficiency ($>30\%$ time reduction) of an L-SLAM system.
Furthermore, by evaluating the environment's degeneracy and adaptively setting the number of good features, our method also works well in non-ideal cases.
Overall, our work presents the following contributions:
\begin{enumerate}
    \item We transform the good feature section in LiDAR-based pose estimation 
    into a problem that preserves the spectral property of information matrices.
    \item We propose and integrate the feature selection method into a multi-LiDAR SLAM system.
    We also propose to evaluate the environment's degeneracy and adaptively change the number of good features online.
    \item We evaluate our approach under extensive experiments with two sensor setups, and computation platforms in terms of accuracy, robustness, and latency. 
\end{enumerate}

\section{Related Work}

Scholarly works on SLAM are extensive. 
Since we focus on the optimal feature selection to improve L-SLAM's efficiency, we review related works on two research topics: feature extraction and selection.

\subsection{Feature Extraction}
Feature extraction is a process to build an informative,  compact, and interpretable representation of raw measurements \cite{guyon2008feature}. 
It has played a crucial role in the front end of many L-SLAM systems to facilitate subsequent tasks.
SuMa and its variants \cite{behley2018efficient,chen2019suma++,chen2020overlapnet} convert point clouds into range images and generated surfel-based maps.
In contrast, LOAM \cite{zhang2014loam} was proposed to extract features on both edge lines and planar surfaces, 
and LEGO-LOAM \cite{shan2018lego} leverages ground features to constrain poses in the vertical direction.
The following approaches apply visual detection \cite{chen2020sloam} 
or directly used dense scanners \cite{bosse2012zebedee,lin2020decentralized} to enhance the feature extraction in noisy or structureless environments.
To decrease the number of features, Zhao et al. \cite{zhou2020roi} presented a probabilistic framework to extract important region of interest by calculating features' densities and distributions. 
This solution enables the removal of dynamic objects and redundant points in busy urban environments.

All of these methods extract features depending on local geometric structures, 
but they have not considered the explicit relationship between the pose estimation 
to select the most useful features.
Our system is built on LOAM's framework in feature selection. 
As a complement to the above appearance-based approaches, 
our solution identifies a set of good features by utilizing motion information. 

\subsection{Feature Selection}
Motion-based feature selection methods have been widely applied in visual SLAM \cite{choudhary2015information,carlone2018attention,zhao2020good,fontan2020information}, 
and many of them are based on the covariance or information matrices 
that capture uncertainties of poses \cite{kaess2009covariance}.
These methods formulate the feature selection as a submatrix selection problem 
and aim to find a subset of features with the objective to preserve the information matrix's spectral attributes. 
Recent works by Carlone et al. \cite{carlone2018attention} and Zhao et al. \cite{zhao2020good} have investigated greedy-based algorithms \cite{mirzasoleiman2015lazier} to solve this NP-hard feature selection problem at a polynomial-time complexity.

A common limitation of Carlone's and Zhao's works is that they assure sufficient features to be available. 
Under this assumption, the pose optimization with a set of good features remains well-conditioned.
However, robots sometimes need to work in degraded environments 
such as textureless regions for cameras and narrow corridors for LiDARs.
Therefore, only utilizing good features with a fixed size becomes degenerate. 
Based on Zhao's feature selection approach, 
our method additionally evaluates environments' degeneracy online, 
which enables us to adaptively change the number of good features to avoid the risk of ill conditions.

\section{Nonlinear Least-Squares Pose Estimation}
\label{sec.nls}

We formulate the pose estimation of a L-SLAM system as an maximum likelihood estimation (MLE) \cite{barfoot2017state}:
\begin{equation}
\label{equ.mle}
\begin{split}
    \hat{\mathbf{x}}_{K}
    &=
    \underset{\mathbf{x}_{K}}{\arg\max}\ 
    p(\mathcal{F}_{K}|\mathbf{x}_{K})
    =
    \underset{\mathbf{x}_{K}}{\arg\min}\ 
    f(\mathbf{x}_{K},\mathcal{F}_{K}),
\end{split}
\end{equation}
where $\mathcal{F}_{K}$ represents the available features at the $K^{th}$ frame, 
$\mathbf{x}_{K}$ is the robot's pose to be optimized, 
and $f(\cdot)$ is the objective function.
Assuming the measurement model to be Gaussian, 
problem \eqref{equ.mle} is solved as an NLS problem:
\begin{equation}
\label{equ.nls}
\begin{split}
    \hat{\mathbf{x}}_{K}
    =
    \underset{\mathbf{x}_{K}}{\arg\min}\ 
    \sum_{i=1}^{N}
    \rho\big(
    \big|\big|
    \mathbf{r}(\mathbf{x}_{K},\mathbf{p}_{Ki})\big|\big|^{2}_{\mathbf{W}_{i}}
    \big),
\end{split}
\end{equation}
where
$\rho(\cdot)$ is the robust loss \cite{bosse2016robust} and
$\mathbf{W}_{i}$ is the covariance matrix.
Problem \eqref{equ.nls} is equivalently rewritten as \cite{barfoot2017state}
\begin{equation}
\label{equ.irls}
    \hat{\mathbf{x}}_{K}
    =
    \underset{\mathbf{x}_{K}}{\arg\min}\ 
    \sum_{i=1}^{N}
    \big|\big|
    \mathbf{r}(\mathbf{x}_{K},\mathbf{p}_{Ki})\big|\big|^{2}_{\bm{\Sigma}_{i}},    
\end{equation}
where $\bm{\Sigma}_{i}^{-1}=\rho'(||\mathbf{r}(\mathbf{x}_{\text{op},K},\mathbf{p}_{Ki})||^{2}_{\mathbf{W}_{i}}) \mathbf{W}_{i}^{-1}$ 
is the alternative covariance matrix 
and $\rho'(\cdot)$ is the derivative of $\rho(\cdot)$.
\eqref{equ.nls} is simplified as an iterative reweighed least-squares problem.
Iterative methods such as Gauss-Newton or Levenberg-Marquardt can offen be used to solve this problem.
These methods locally linearize the objective function by computing the Jacobian w.r.t. $\mathbf{x}_{K}$ as $\mathbf{J}=\partial f/\partial\mathbf{x}_{K}$.
Given an initial guess, $\mathbf{x}_{K}$ is iteratively optimized by usage of $\mathbf{J}$ until convergence to find an optimum.

At the final iteration, the least-squares covariance of the state is calculated as $\bm{\Xi}=\bm{\Lambda}^{-1}$ \cite{censi2007accurate}, 
where $\bm{\Lambda}=\mathbf{J}^{\top}\mathbf{J}$ is called the \textit{information matrix}.
This equation reveals the relationship between the pose covariance and information matrix.
Generally, exploiting plentiful measurements in optimization should minimize poses' uncertainties.
This explains why SLAM systems commonly use all available features.

This paper focuses on low-latency applications in which the speed is highly prioritized.
It requires us to utilize only a subset of features to accelerate the algorithm.
As suggested in \cite{zhao2020good}, we can check whether a feature is selected or not by 
comparing the gains in spectrum of $\bm{\Lambda}$.
The word: ``spectrum'' denotes the set of eigenvalues of a matrix \cite{golub2013matrix}.

\section{Methodology}
\label{sec.methodology}

This section first formulates the good feature selection problem and introduces a metric to guide the selection. 
We apply the stochastic-greedy method, which combines the random sampling procedure, to solve this problem in real time.
We then extend this algorithm to achieve efficient feature selection.
Finally, we propose to evaluate the environment's degeneracy online to avoid ill-conditioning estimation.

\subsection{Problem Formulation}
\label{sec.gf_problem_formulation}
We denote $N=|\mathcal{F}_{K}|$ as the number of all available features, 
$M$ as the maximum number of selected features, and $\mathcal{S}_{K}$ as the good feature set. 
We denote $f(\cdot)$ as the metric to quantify the spectral attribute of a matrix. 
We formute the feature selection problem under a cardinality constraint as
\begin{equation}
\label{equ.gf-problem}
    \underset{\mathcal{S}_{K} \subset\mathcal{F}_{K}}
    {\arg\max}\ 
    f\big[\bm{\Lambda}(\mathcal{S}_{K})\big]
    \ \ \ 
    \text{subject\ to}\ |\mathcal{S}_{K}|\leq M,
\end{equation}
where $\bm{\Lambda}(\mathcal{S}_{K})$ is the information matrix on the good feature set.
There are several options to define $f(\bm{\Lambda})$:
$\text{tr}(\bm{\Lambda})$ the trace \cite{summers2015submodularity}, 
$\lambda_{\min}(\bm{\Lambda})$ the minimum eigenvalue \cite{carlone2018attention}, 
and $\log\det(\bm{\Lambda})$ the log determinant \cite{zhao2020good}.
%
%
%
%

Since problem \eqref{equ.gf-problem} is NP-hard, we cannot find efficient algorithms to obtain the optimal subset for real-time applications.
Fortunately, all these metrics are submodular and monotone increasing \cite{mirzasoleiman2015lazier}, allowing the solution to be approximate via greedy methods with a performance guarantee. 
Zhao et al. \cite{zhao2020good} experimented with these metrics in bundle adjustment, 
where the log determinant was demonstrated to have the lowest pose error and computation time.
We thus choose the $\log\det(\bm{\Lambda})$ option as our metric.

\subsection{Stochastic Greedy Algorithm}
The class of greedy methods has been studied to solve problem \eqref{equ.gf-problem}. 
Here, we introduce the stochastic-greedy algorithm \cite{mirzasoleiman2015lazier}, which applies randomized acceleration to avoid brute-force search.
The idea is simple: at each round, the current best feature is picked up by examining all elements from a random subset. 
This is different from the simple greedy approach, which has to search the whole set.
We define the size of the random subset as $\frac{N}{M}\log(\frac{1}{\epsilon})$, where $\epsilon$ is the \textit{decay factor}.
The time complexity is $O(N\log(\frac{1}{\epsilon}))$, which is independent of $M$.
The stochastic-greedy algorithm has been proved to have near-optimal performance in \cite{mirzasoleiman2015lazier}:

\textbf{Theorem 1:} Let $f(\cdot)$ be the non-negative, monotone, and submodular function. 
Setting the size of the random subset as $\frac{N}{M}\log(\frac{1}{\epsilon})$. 
Denote $\mathcal{S}_{K}^{*}$ as the optimal set and $\mathcal{S}_{K}^{\#}$ the result found by the stochastic-greedy algorithm. 
$\mathcal{S}_{K}^{\#}$ enjoys the approximation guarantees in expectation:
\begin{equation}
\label{equ.theorem1}
E
\Big(
f\big[
\bm{\Lambda}(\mathcal{S}_{K}^{\#})
\big]
\Big)
\geq
\underbrace{(1-1/e-\epsilon)}_{\text{expected\ ratio}}
f\big[
\bm{\Lambda}(\mathcal{S}_{K}^{*})
\big].
\end{equation}

\subsection{Good Feature Selection for Pose Estimation}
\label{sec.gf-lslam}
\begin{algorithm}[t]
    \caption{Stochastic Greedy-Based Good Feature Selection for NLS Pose Estimation}
    \label{alg.sto-greedy}        
    \LinesNumbered
    \KwIn{$f(\cdot)\triangleq\log\det(\cdot)$, 
    $M$, $\epsilon$,  
    $\mathcal{M}_{K}$, 
    $\mathcal{F}_{K}$, $\mathbf{x}_{K}$;}
    \KwOut{good feature set $\mathcal{S}_{K}^{\#}$;}
    Initialize the set $\mathcal{S}_{K}^{\#} = \emptyset$;\\
    \While{$|\mathcal{S}_{K}^{\#}|<M$ and $t_{comp}<t_{max}$}
    {
        $\mathcal{R}\leftarrow$ the random subset is obtained by sampling $\frac{N}{M}\log(\frac{1}{\epsilon})$ 
        random elements from $\mathcal{F}_{K}\backslash\mathcal{S}^{\#}_{K}$;\\
        \ForEach{$\mathbf{p}_{i}\in\mathcal{R}$}
        {
            Search the correspondence from $\mathcal{M}_{K}$;\\
            \If{the correspondence is found}
            {
                Compute the residual $\mathbf{r}_{i}(\check{\mathbf{x}}_{K})$;\\
                Compute     
                $\bm{\Lambda}_{i}
                =
                \mathbf{J}^{\top}_{i}\bm{\Sigma_{i}}^{-1}\mathbf{J}_{i}$ w.r.t. $\mathbf{p}_{i}$ where $\mathbf{J}_{i}$ is the Jacobian of $\mathbf{r}_{i}(\cdot)$;\\
            }           
            \Else
            {
                $\mathcal{R}\leftarrow\mathcal{R}\backslash\{\mathbf{p}_{i}\}$;\ \ \  $\mathcal{F}_{K}\leftarrow\mathcal{F}_{K}\backslash\{\mathbf{p}_{i}\}$;\\
            }           
        }
        $i^{*}\leftarrow\arg\max_{\mathbf{p}_{i}\in\mathcal{R}}\log\det\big[\bm{\Lambda}(\mathcal{S}_{K}^{\#}) + \bm{\Lambda}_{i}\big]$;\\
        $\bm{\Lambda}(\mathcal{S}_{K}^{\#})\leftarrow\bm{\Lambda}(\mathcal{S}_{K}^{\#}) + \bm{\Lambda}_{i^{*}}$;\\      
        $\mathcal{S}_{K}^{\#}\leftarrow\mathcal{S}_{K}^{\#}\cup\{\mathbf{p}_{i^{*}}\}$;\ \ \  $\mathcal{F}_{K}\leftarrow\mathcal{F}_{K}\backslash\{\mathbf{p}_{i^{*}}\}$;\\
    }
\end{algorithm}


Based on the theoretical results, we employ the stochastic-greedy to achieve the feature selection for pose estimation.
The detailed pipeline is summarized in Algorithm \ref{alg.sto-greedy}.

\begin{enumerate}
    \item Overview:
    The algorithm starts with the $\log\det(\cdot)$ metric, 
    the number of good features $M$, 
    the \textit{decay factor} $\epsilon$,   
    the map in the reference frame $\mathcal{M}_{K}$, 
    the set of all available features in the current frame $\mathcal{F}_{K}$, 
    and
    the initial pose $\mathbf{x}_{K}$.
    It produces the good feature set $\mathcal{S}_{K}^{\#}$.
    
    \item Line $2$: The loop is exited if one of the following conditions is satisfied: $M$ good features are found or the computation time exceeds $t_{max}$.
    The second condition limits the cost of finding good features.
    Since $\log\det(\cdot)$ is submodular with diminishing returns, early termination does not induce much information loss.
    
    \item Lines $4$--$10$: The correspondence of each feature in the random subset $\mathcal{R}$ is found
    from $\mathcal{M}_{K}$. 
    The residual is then computed. 
    If a feature has already been visited at previous iterations, we skip these steps.
    
    \item Lines $11$--$13$: The feature which leads to the maximum enhancement of the objective is selected. 
    After that, the information matrix $\bm{\Lambda}(\mathcal{S}^{\#}_{K})$, $\mathcal{S}^{\#}_{K}$, and $\mathcal{F}_{K}$ are updated.
\end{enumerate}
      
Furthermore, the process of feature selection implicitly performs outlier rejection: outliers are penalized by the robust loss in \eqref{equ.irls} with relatively small weights.
They contribute less to $\bm{\Lambda}$ than standard features and will be selected with a low probability.
Therefore, selecting good features might reduce the biases between estimates and the ground truth.

\subsection{Setting the Number of Good Features}
\label{sec.gf-subset-size}
Setting a proper size for the good feature set is essential to the system.
Previous methods manually set $M$ as a constant value ($M=100$ \cite{zhao2020good}) or a fixed ratio of all features ($M=0.5N$ \cite{carlone2018attention}). 
These schemes are feasible if sufficient features are always available.
But if a robot has to work in non-ideal scenarios such as textureless walls or narrow corridors, utilizing a small set of features is not reliable.
On the other hand, if we change the hard-coded number $M$ in a specific situation, it will inevitably increase the cost of deploying and maintaining a SLAM system on real platforms.

It would be better if $M$ were adaptively changed by evaluating the degeneracy online.
Inspired by \cite{zhang2016degeneracy}, the magnitude of the degeneracy can be quantified by a \textit{factor} $\lambda$.
Differently, we define the factor using the log determinant metric as  $\lambda=\log\det\bm{\Lambda}(\mathcal{F}_{K})$.
Computing the information matrix on the full feature set is time-consuming.
Since the robot performs a continuous movement in an environment, it is enough to compute $\lambda$ at every time interval ($1s\sim2s$). 

\begin{figure}[]
	\centering
	\includegraphics[width=0.42\textwidth]{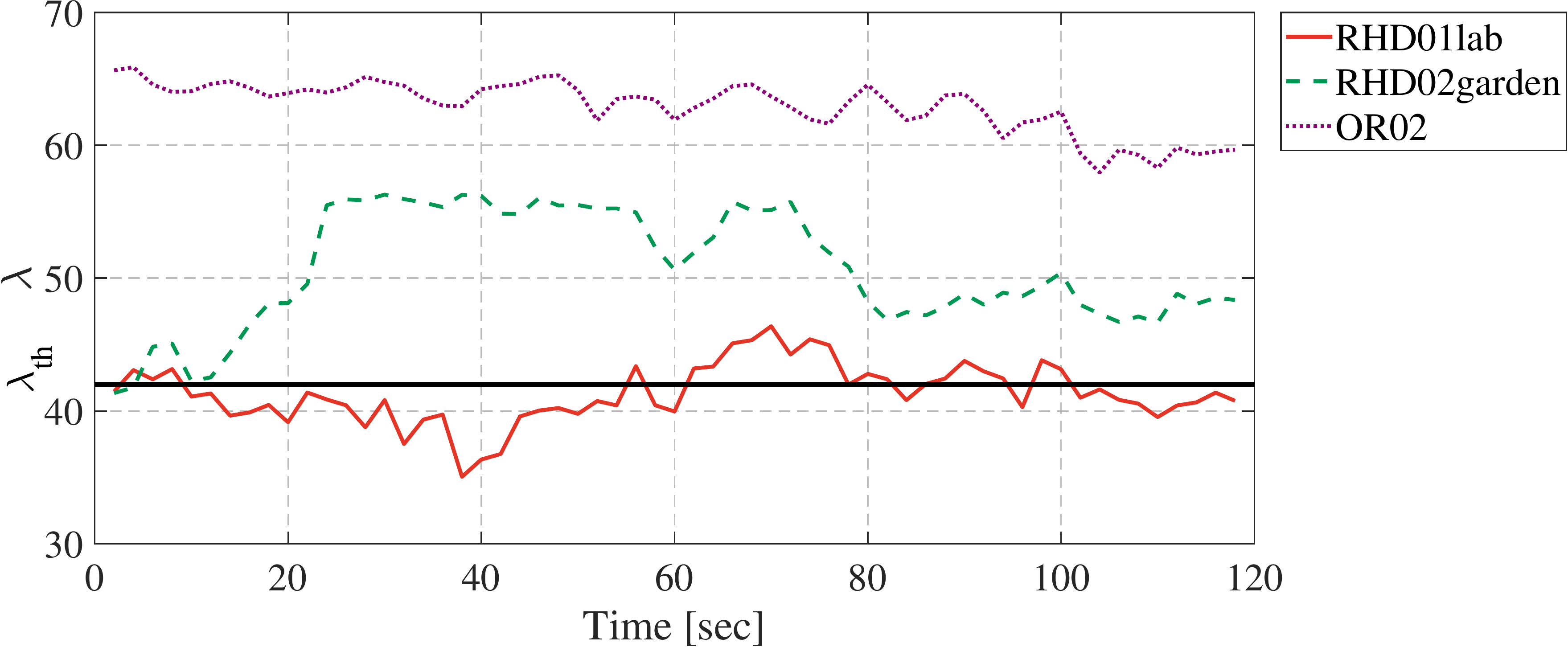}
	\caption{The value of the \textit{degeneracy factor} $\lambda$ on different sequences.}
	\label{fig.logdet_lambda}
\end{figure}

Fig. \ref{fig.logdet_lambda} plots the values of $\lambda$ on different sequences.
\textit{RHD01lab} contains several degenerate scenarios, while other sequences do not (see Section \ref{sec.exp-per-slam}). 
Therefore, we empirically set $\lambda_{th}=42$.
If $\lambda\geq\lambda_{th}$, we select $20\%$ of features from the full set as the good features (i.e., $M=0.2N$). 
Otherwise, we use $80\%$ of features from the set.


\section{GF-Enhanced Multi-LiDAR SLAM System}

\begin{figure}[]
	\centering
	\subfigure[Full feature set ($4693$ points).]
	{\label{fig.full}\centering\includegraphics[width=0.23\textwidth]{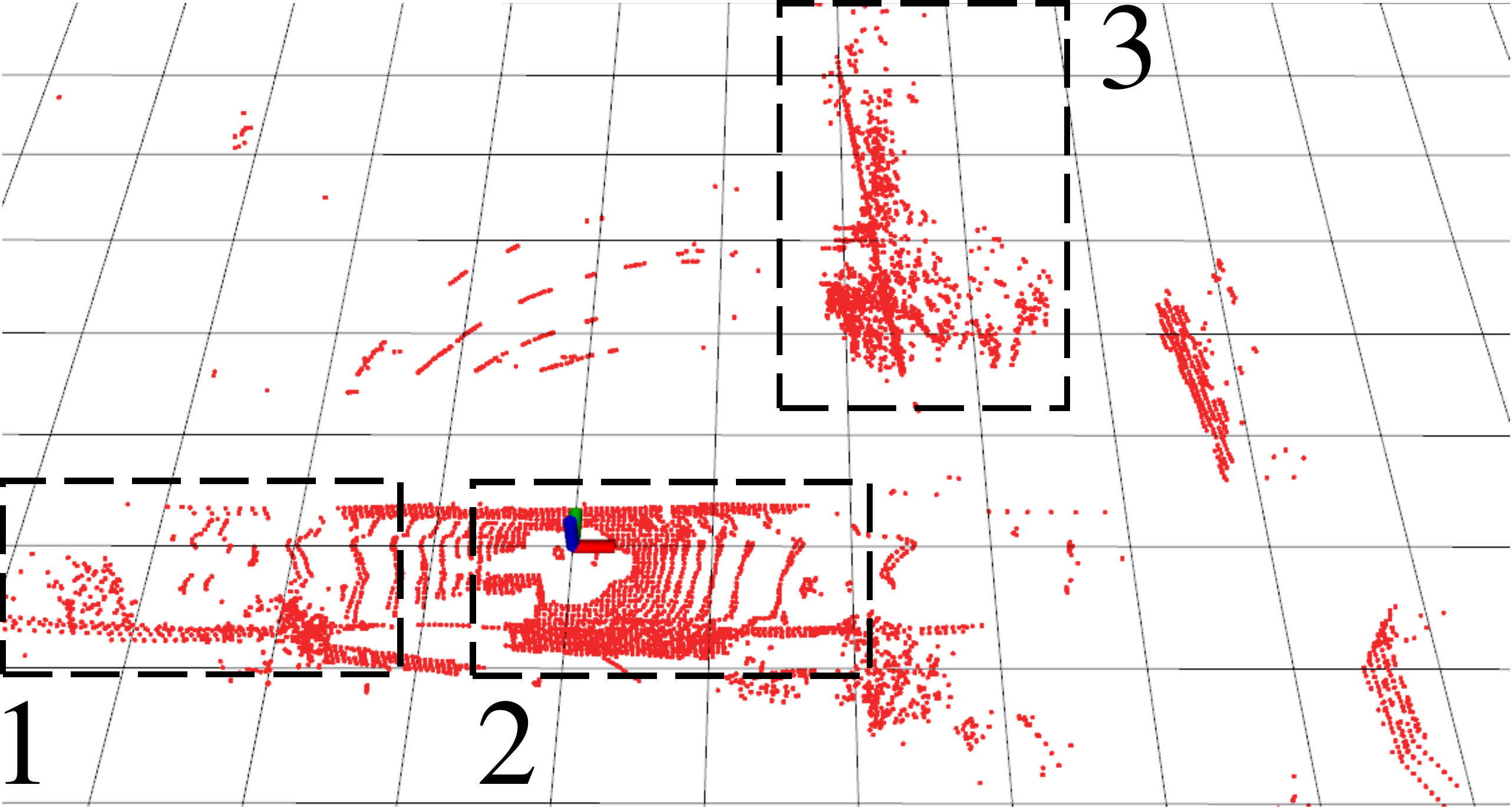}}
	\subfigure[Good features ($929$ points).]
	{\label{fig.gf}\centering\includegraphics[width=0.23\textwidth]{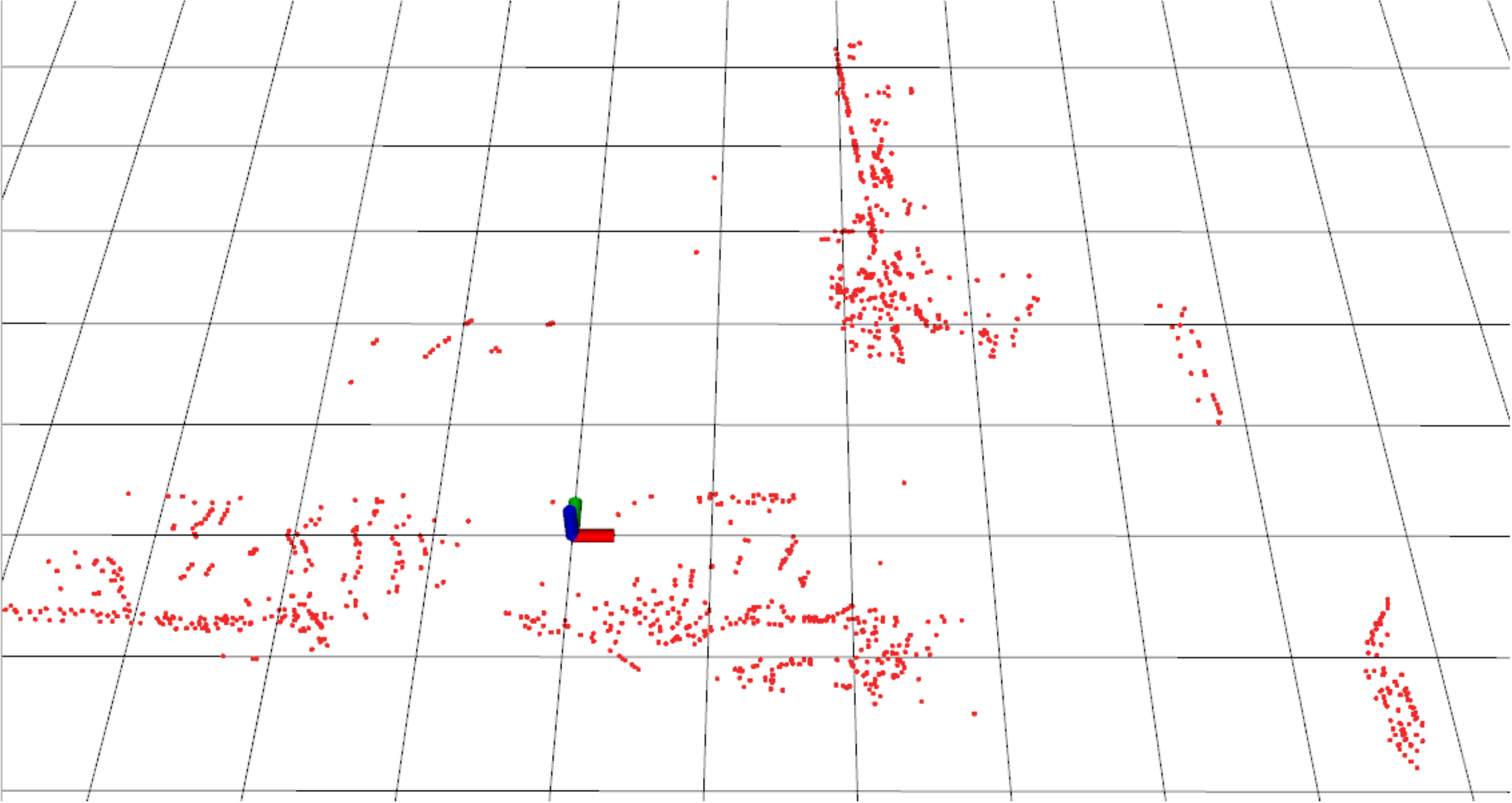}}
	\caption{A qualitative example of good features selected by our greedy-based feature selection algorithm. This method pick up points on objects which provide strong geometric constraints, as indicated in the region 1 and 3. 
	Points on the ground, which occupy around $50\%$ of the full feature set, are indicated by the region 2. Since they only constrain poses on the $z-$ axis, our method only selected them with a small number.
	This is the main difference from the fully random sampling method (see Section \ref{sec.exp-per-gf}).}
	\label{fig.full_gf}	
\end{figure}  

\begin{figure}[]
    \centering
    \includegraphics[width=0.48\textwidth]{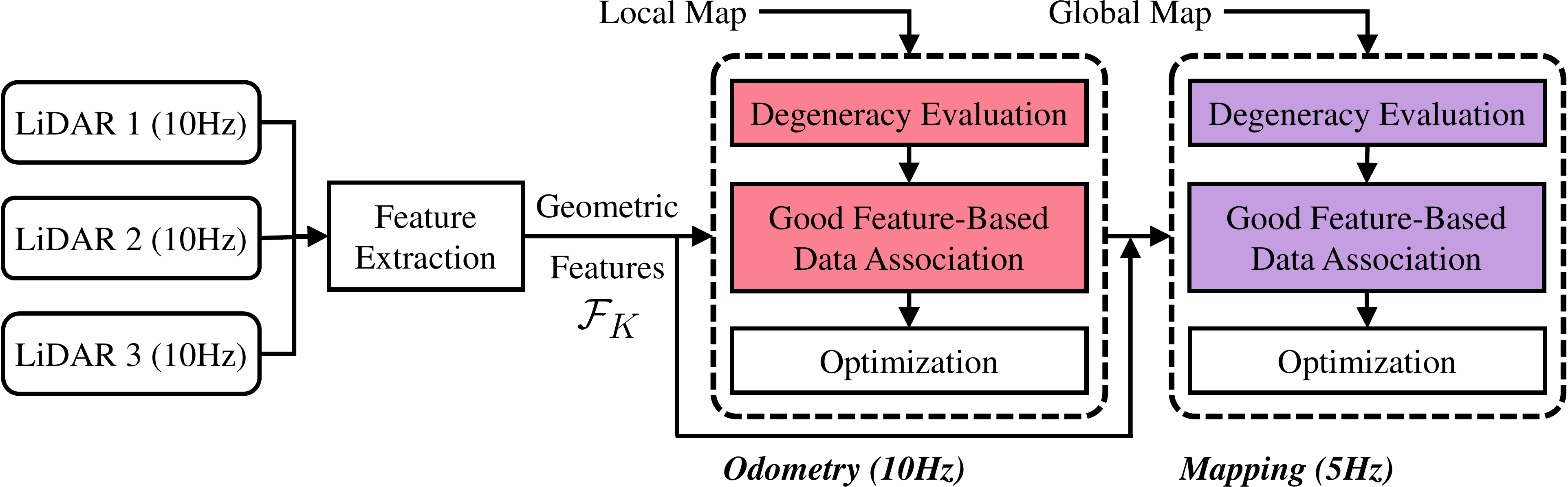}  
    \caption{Block diagram of the pipeline of M-LOAM-gf.}
    \label{fig.system_pipeline}
\end{figure}

The proposed feature selection method has been verified in an L-SLAM  system called M-LOAM \cite{jiao2020robust}.
To distinguish it from the original system, the enhanced system is denoted by \textbf{M-LOAM-gf}. 
M-LOAM-gf solves SLAM with multiple LiDARs by two algorithms: \textit{odometry} and \textit{mapping}.
Generally, these algorithms are designed to estimate poses in a coarse-to-fine fashion.
Since they similarly formulate the NLS problem for pose estimation, the feature selection can be applied to both.
Fig. \ref{fig.system_pipeline} illustrates the overall structure of M-LOAM-gf. 
Note that loop closure is not included.

We give real definitions to the feature set $\mathcal{F}_{K}$.
We extract features located on local edges and planar surfaces from the LiDARs' raw measurements.
According to the points' curvatures, we select a set of edge and planar features to form $\mathcal{F}_{K}$.
The next step is to match features between the reference frame and the robot's current frame. 
In both \textit{odometry} and \textit{mapping}, we use the feature map in the reference frame to associate data with $\mathcal{F}_{K}$.
The only difference is the scale. The local map in \textit{odometry} is built within a small time interval ($<0.5s$), while the global map in \textit{mapping} is constructed using all features in keyframes.
For convenience, we use $\mathcal{M}_{K}$ to denote both the local and global maps.

With the found correspondences, we can optimize the relative transformation by minimizing the sum of all errors.
The good feature selection algorithm enables M-LOAM-gf to select only a set of features in optimization while preserving the spectrum of the information matrix.
An example of good features are shown in Fig. \ref{fig.full_gf}.
After obtaining the good feature set $\mathcal{S}^{\#}_{K}$ in Section \ref{sec.gf-lslam}, the objective function is written as
\begin{equation}
\label{nls-with-gf}
    \hat{\mathbf{x}}_{K}
    =
    \underset{\mathbf{x}_{K}}{\arg \min}
    \sum_{\mathbf{p}\in\mathcal{S}^{\#}_{K}}^{} 
    \big|\big|
        \mathbf{r}(\mathbf{x}_{K},\mathbf{p})
    \big|\big|_{\bm{\Sigma}_{\mathbf{p}}}^{2},
\end{equation}
for which the expression of the residuals and Jacobians is detailed in our supplementary material \cite{jiao2020supplementary}.

\section{Experiment}
\label{sec.experiment}

\begin{table}[]
	\centering
	\caption{Mean and std of $\log\det\bm{\Lambda}(\mathcal{S}_{K}^{\#})$ on RHD and OR sequences.}
	\renewcommand\arraystretch{1.25}
	\renewcommand\tabcolsep{7pt}
	\begin{tabular}{|c|cc|c|}
		\hline
		Sequence & Greedy & Rnd & Full \\ 
		\hline		
		\textit{RHD01lab} & $\bm{35.1}\pm\bm{2.3}$ & $30.3\pm2.7$ & $\darkgraytext{39.3}\pm\darkgraytext{2.5}$ \\
		\textit{RHD02garden} & $\bm{41.4}\pm\bm{4.0}$ & $39.7\pm5.0$ & $\darkgraytext{48.2}\pm\darkgraytext{5.1}$ \\
		\hline
		\textit{OR01} & $52.6\pm\bm{1.0}$ & $\bm{52.7}\pm1.6$ & $\darkgraytext{61.8}\pm\darkgraytext{1.7}$ \\
		\textit{OR02} & $50.8\pm\bm{0.9}$ & $\bm{51.8}\pm2.2$& $\darkgraytext{60.1}\pm\darkgraytext{2.1}$ \\
		\textit{OR03} & $\bm{51.7}\pm\bm{1.4}$ & $49.9\pm2.6$& $\darkgraytext{58.3}\pm\darkgraytext{2.4}$ \\
		\textit{OR04} & $\bm{51.6}\pm\bm{1.2}$ & $50.3\pm2.2$& $\darkgraytext{58.7}\pm\darkgraytext{2.1}$ \\
		\textit{OR05} & $\bm{52.5}\pm\bm{1.7}$ & $51.9\pm2.7$ & $\darkgraytext{59.3}\pm\darkgraytext{2.4}$ \\
		\hline		
		Average & $\bm{47.96}$ & $46.66$ & $\darkgraytext{55.10}$ \\
		\hline
	\end{tabular}
	\label{tab.per-gf}
\end{table}

\begin{figure*}[]
	\centering
	\subfigure[The real handheld device (RHD).]
	{\label{fig.exp_rhd}\centering\includegraphics[width=0.28\textwidth]{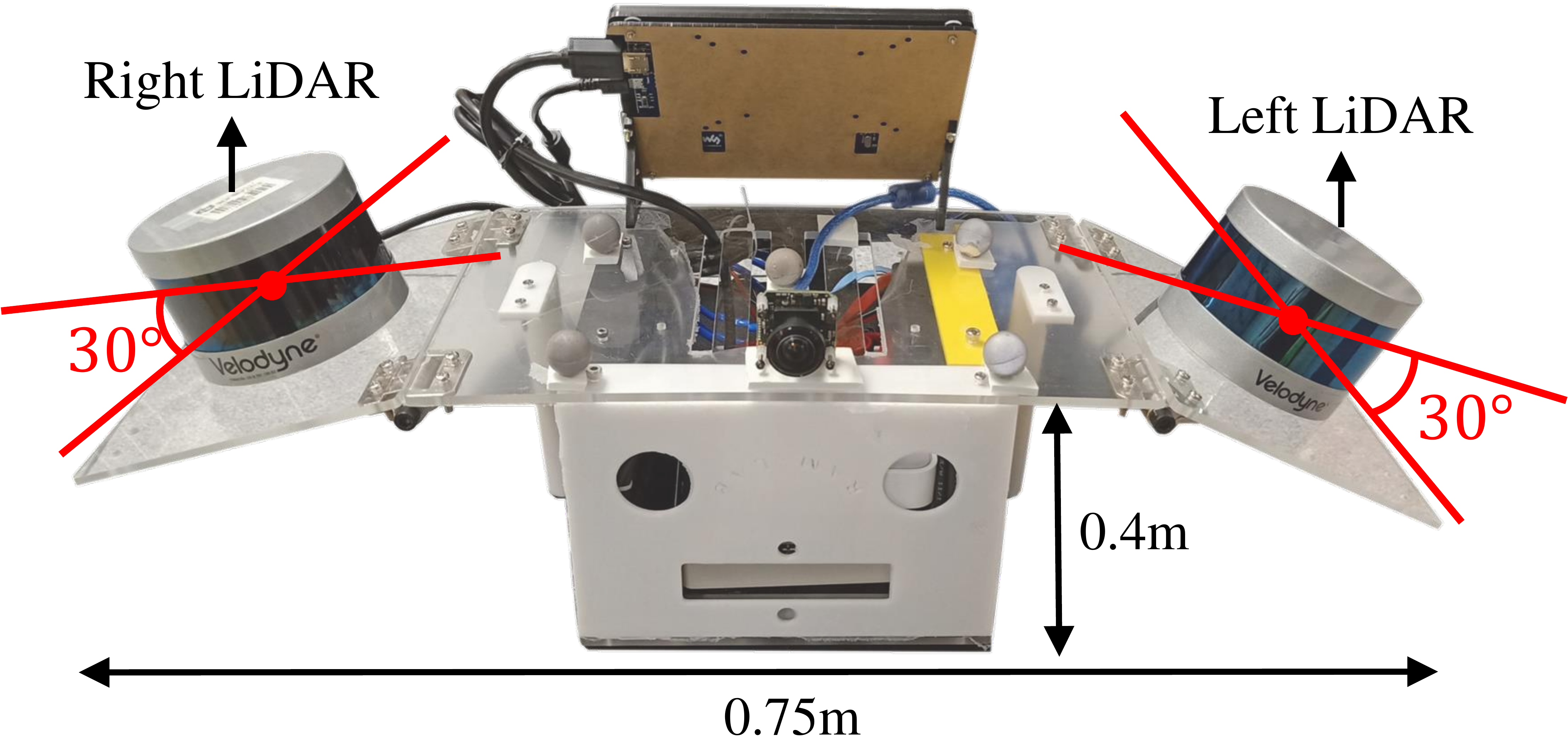}}
	\hspace{0.05cm}
	\subfigure[Results on the \textit{RHD01lab}.]
	{\label{fig.exp_rhd02traj}\centering\includegraphics[width=0.38\textwidth]{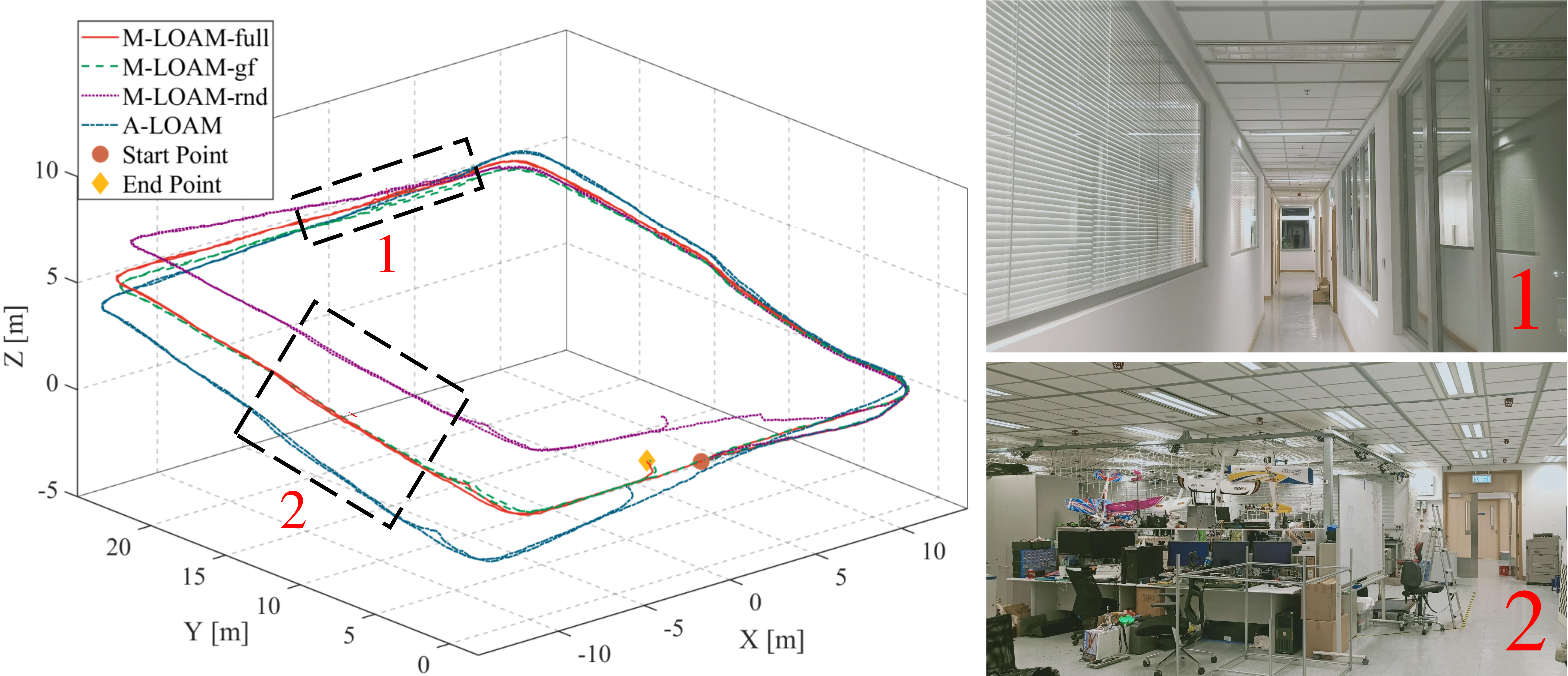}}
	\hspace{0.15cm}	
	\subfigure[Results on the \textit{RHD02garden}.]
	{\label{fig.exp_rhd03traj}\centering\includegraphics[width=0.27\textwidth]{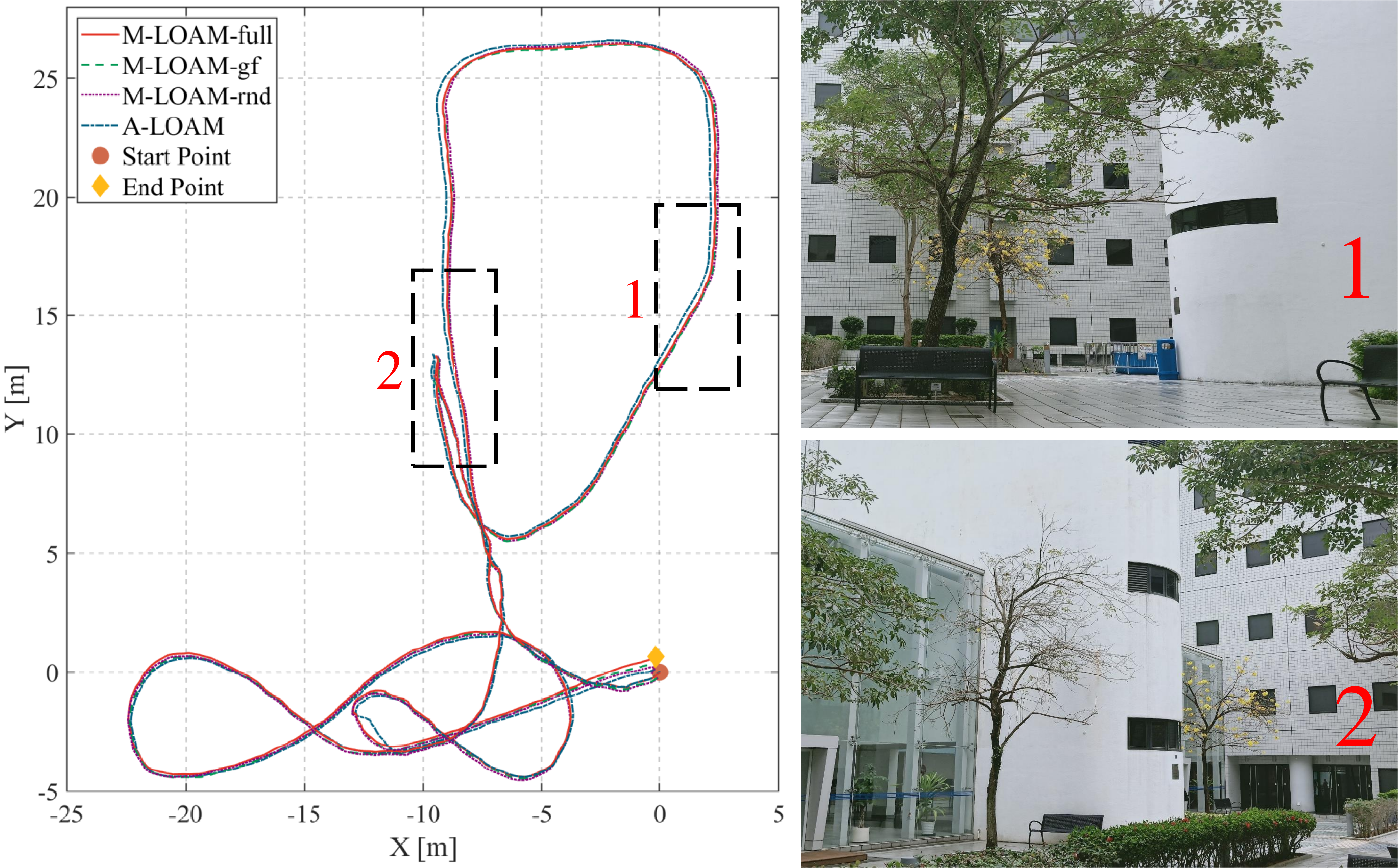}}
	\caption{Estimated trajectories of different methods and the scene image on two \textit{RHD} sequences. }
	\label{fig.rhd_traj}
\end{figure*}  

\begin{figure*}[]
	\centering
	\includegraphics[width=0.96\textwidth]{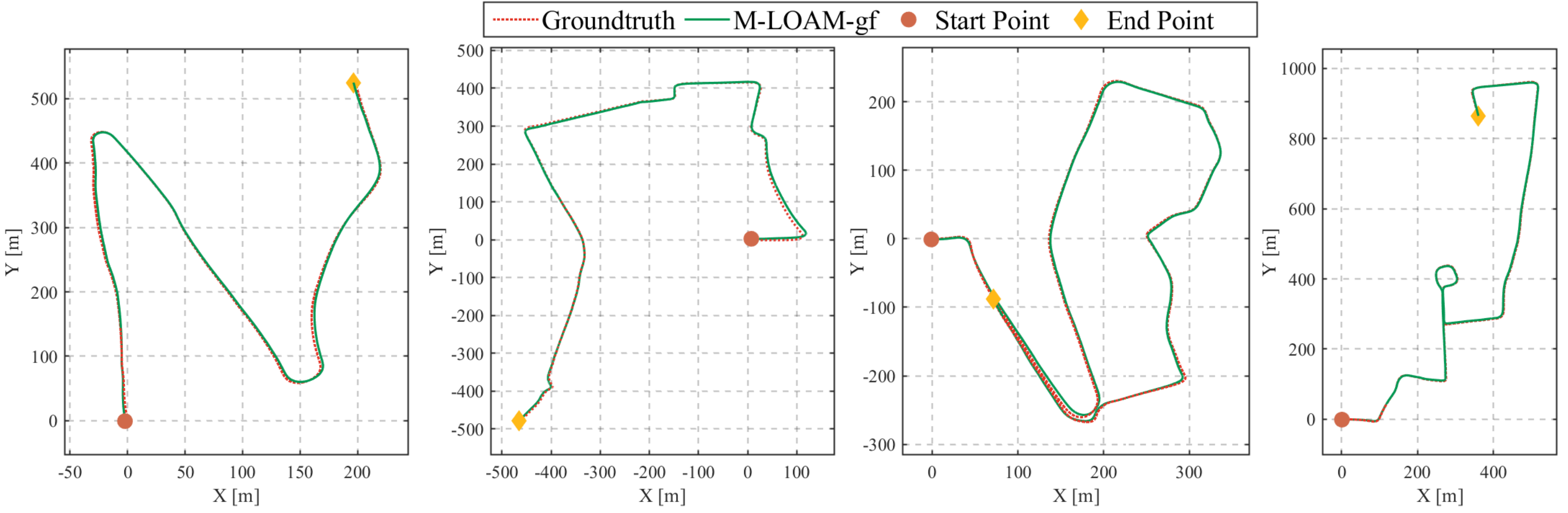}
	\caption{M-LOAM-gf's trajectories on \textit{OR01}, \textit{OR03}, \textit{OR04}, and \textit{OR05} from the Oxford Robocar dataset \cite{barnes2019oxford} are aligned with the ground truth .}
	\label{fig.exp_or_others_trajectory}
\end{figure*}

\begin{figure}[]
	\centering
	{\label{fig.exp_or_02_pc}\centering\includegraphics[width=0.23\textwidth]{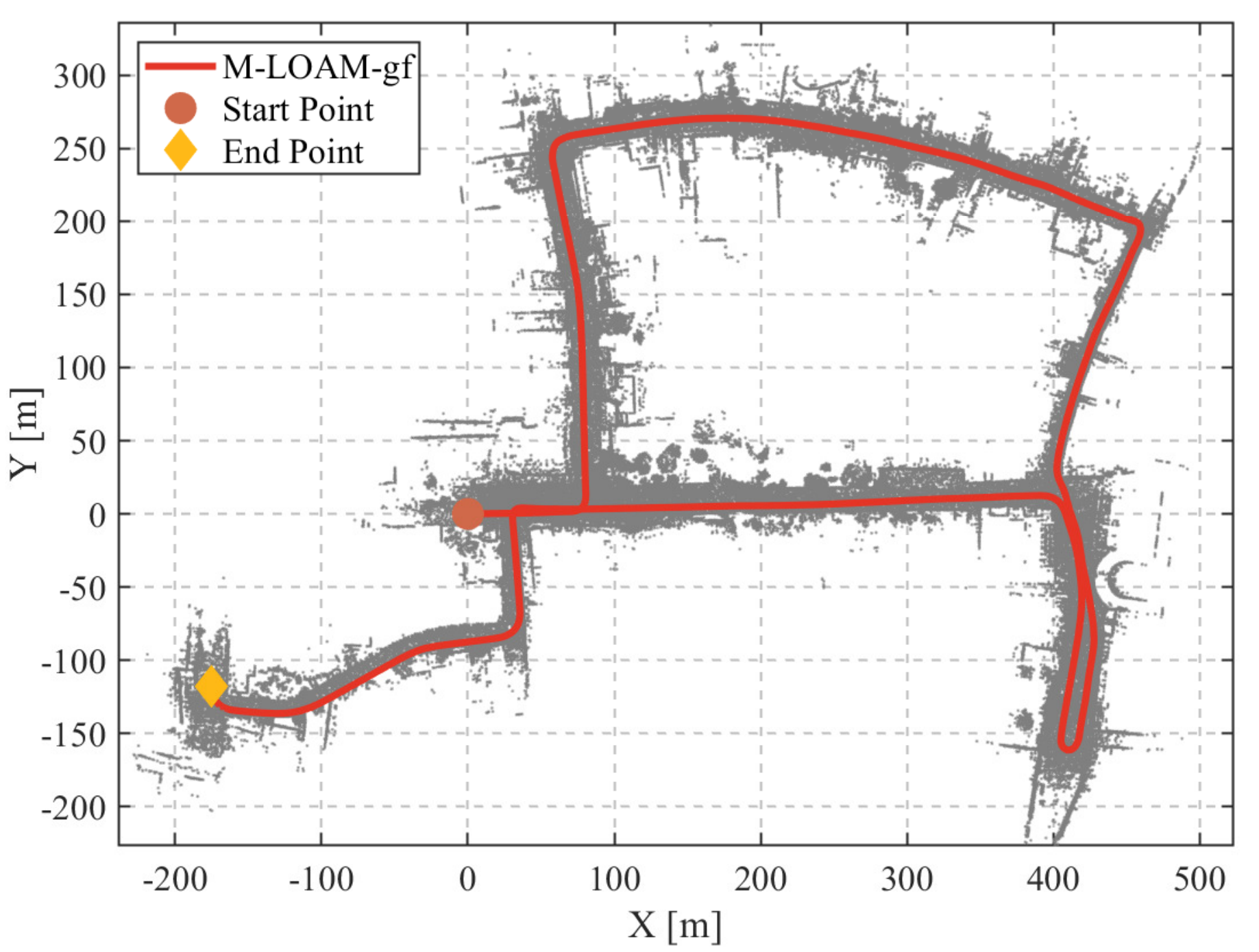}}
	{\label{fig.exp_or_02_traj}\centering\includegraphics[width=0.246\textwidth]{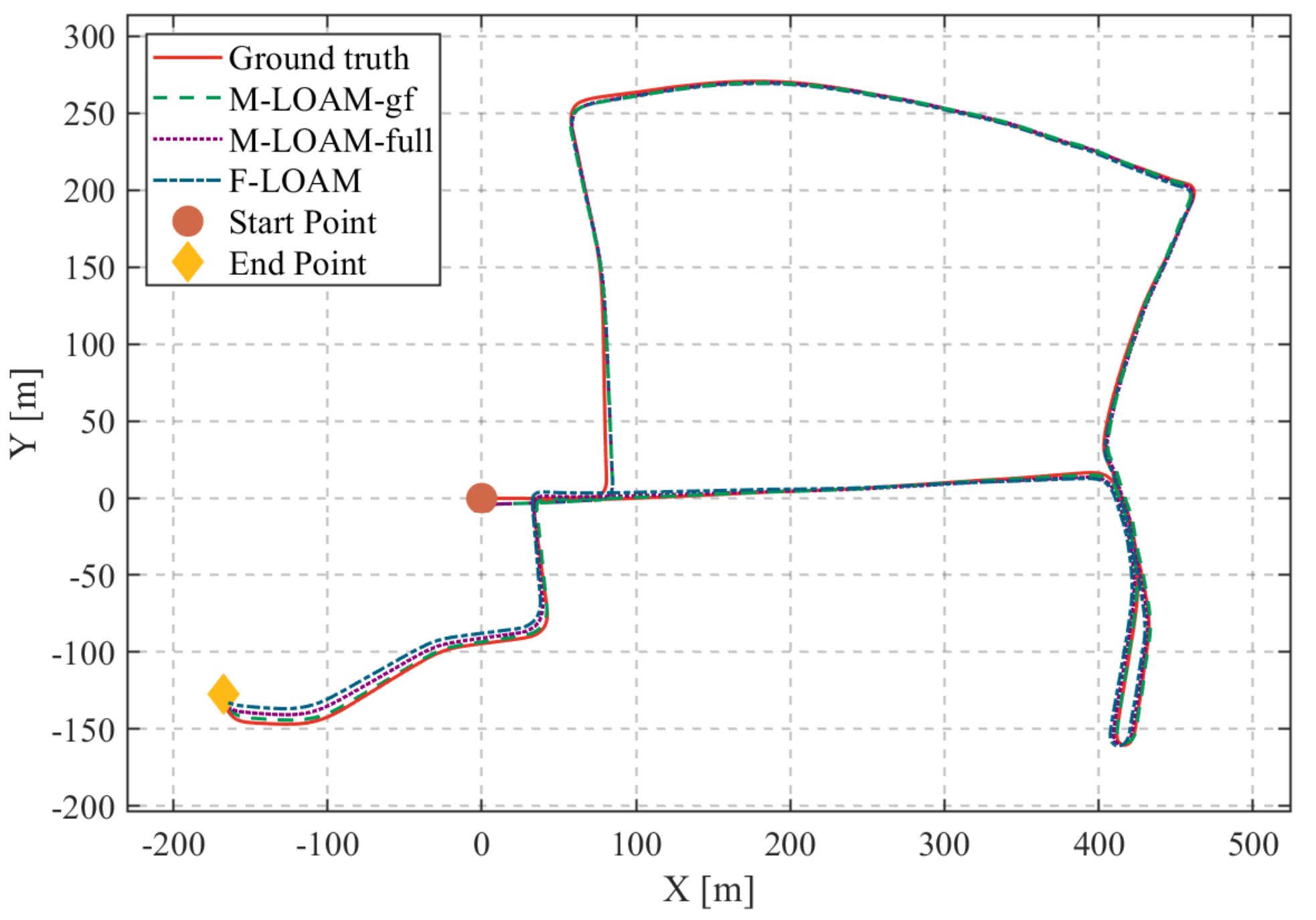}}
	\caption{Results on the \textit{OR02}. (left) Map and M-LOAM-gf's path. (right) Estimated trajectories of different methods are aligned with the ground truth.}
	\label{fig.exp_or02}
	\vspace{-0.2cm}
\end{figure}

We evaluate the performance of M-LOAM-gf on real-world experiments.
First, we validate the stochastic-greedy-based feature selection.
Second, we demonstrate the localization accuracy of M-LOAM-gf in various scenarios covering indoor environments and outdoor urban roads with two multi-LiDAR setups.
Three SOTA L-SLAM systems are compared.
We also study M-LOAM-gf's latency on an on-board processor with limited computation resources.

\subsection{Implementation Details}
\label{sec.exp-implement}

We use the PCL library \cite{rusu20113d} to process point clouds and the Ceres Solver \cite{agarwal2015others} to solve the NLS problems. 
Our method is tested on sequences collected with two platforms:
\begin{itemize}
	\item \textbf{Real Handheld Device (RHD)} is made for indoor tests and shown in Fig. \ref{fig.exp_rhd}. 
	It is installed with two VLP-16\footnote{\url{https://velodynelidar.com/products/puck}}.
	We held this device to collect two sequences (\textit{RHD01lab} and \textit{RHD02garden}) with an average speed of $2m/s$.

	\item \textbf{Oxford Robocar (OR)} \cite{barnes2019oxford} is a vehicle equipped with two HDL-32E\footnote{\url{https://velodynelidar.com/products/hdl-32e}}. 
	Datasets were recorded by driving the car on urban roads at an average speed of $10m/s$. $32$ repeated traversals of a $9\ km$ route were collected.
	Ground-truth poses in $SE(2)$ are available.
	We select one sequence lasting $34$ minutes and split it into $5$ sequences named \textit{OR01}--\textit{OR05} for evaluation.

\end{itemize}



\subsection{Validation on Good Feature Selection}
\label{sec.exp-per-gf}



This section validates that the greedy algorithm selects a set of valuable features with a large $\log\det\bm{\Lambda}(\mathcal{S}^{\#}_{K})$.
Our stochastic-greedy method (label: \textbf{greedy}) is compared with the fully randomized selection method (label: \textbf{rnd}).
Table \ref{tab.per-gf} reports the means and standard deviations (std) of $\log\det\bm{\Lambda}(\mathcal{S}^{\#}_{K})$.
The values with the full feature set (label: \textbf{full}) are provided for reference.
On \textit{OR01}--\textit{OR02}, the greedy method gains a smaller objective than the rnd method.
This is reasonable since the greedy algorithm cannot always achieve the best performance according to \eqref{equ.theorem1}.
By considering the std and the larger means on most sequences, we conclude that the greedy method outperforms the rnd method.

\begin{table*}[]
	\centering
	\caption{Translational ATE \cite{zhang2018tutorial} on OR sequences (the two best results are marked as bold text).}
	\renewcommand\arraystretch{1.25}
	\renewcommand\tabcolsep{5pt}
	\begin{tabular}{ccccccccc}
		\hline
		\toprule[0.03cm]
		& Seq. & Dimension & M-LOAM-gf & M-LOAM-rnd & M-LOAM-full & A-LOAM & F-LOAM & LEGO-LOAM \\ 
		\hline
		\toprule[0.03cm]
		\multirow{5}{*}{\rotatebox[]{90}{$\textbf{RMSE}_{\mathbf{t}}[m]$}}
		& \textit{OR01} & $525m\times252m$ & $\bm{1.986}\pm0.013$ & $2.324\pm0.028$ & $2.449\pm0.014$ & $4.504\pm0.055$ & $\bm{2.174}\pm0.007$ & $4.334\pm0.000$  \\
		& \textit{OR02} & $629m\times431m$ & $\bm{2.473}\pm0.151$ & $\bm{3.107}\pm0.104$ & $3.120\pm0.076$ & $7.535\pm0.410$ & $3.709\pm0.044$ & $4.369\pm0.083$  \\
		& \textit{OR03} & $573m\times896m$ & $\bm{1.955}\pm0.105$ & $2.829\pm0.158$ & $3.063\pm0.062$ & $12.995\pm0.006$ & $\bm{2.019}\pm0.024$ & $5.440\pm0.102$ \\
		& \textit{OR04} & $337m\times498m$ & $\bm{1.720}\pm0.047$ & $2.242\pm0.084$ & $2.239\pm0.024$ & $\bm{1.913}\pm0.057$ & $2.238\pm0.062$ & $6.224\pm0.021$  \\
		& \textit{OR05} & $517m\times968m$ & $\bm{1.719}\pm0.027$ & $\bm{2.387}\pm0.432$ & $2.567\pm0.487$ & $6.446\pm0.015$ & $2.469\pm0.061$ & $7.248\pm0.260$  \\
		\toprule[0.01cm]
		& Average & & $\bm{1.971}$ & $2.578$ & $2.688$ & $6.679$ & $\bm{2.522}$ & $5.523$ \\
		\toprule[0.03cm]
	\end{tabular}
	\label{tab.or_result}
	\vspace{-0.2cm}
\end{table*}

\subsection{Performance of SLAM}
\label{sec.exp-per-slam}

We compare the accuracy, robustness, and latency of M-LOAM-gf with 
several baseline methods:
\begin{itemize}
    \item \textbf{M-LOAM-rnd} is the variant of M-LOAM with the rnd feature selection module in mapping.
    
    \item \textbf{M-LOAM-full} is the original M-LOAM that uses the full feature set in mapping.
    
    \item \textbf{A-LOAM}\footnote{\url{https://github.com/HKUST-Aerial-Robotics/A-LOAM}}, \textbf{F-LOAM}\footnote{\url{https://github.com/wh200720041/floam}}, and \textbf{LEGO-LOAM} \cite{shan2018lego} are three SOTA, open-source L-SLAM systems. 
    All of them are the improved versions of LOAM \cite{zhang2014loam}. 
    
\end{itemize}

The \textit{odometry} and \textit{mapping} of all methods run at $10$ Hz and $5$ Hz. 
For a fair comparison, the loop closure modules in some baselines are deactivated.
The resolutions of the voxel filter \cite{rusu20113d} on the edge and planar features are $0.2m$ and $0.4m$.

\subsubsection{Qualitative Comparison}
\label{sec.exp-per-qualitative}
We first test our method on RHD sequences.
\textit{RHD01lab} is recorded by moving around an office space, 
in which several scenes provide only poor geometric constraints. Fig. \ref{fig.exp_rhd02traj} indicates two examples. Scene 1 is a long and narrow corridor, which is a typical degenerate environment \cite{ye2019tightly}. Scene 2 is an indoor office, providing well-conditioned constraints. 
M-LOAM-gf successfully tracks robot poses due to its capability in evaluating the environment's degeneracy. 
M-LOAM-rnd has a sudden drift in scene 1 since using only $20\%$ of features cannot constrain the poses. 
A-LOAM also fails because it cannot model the uncertainty in mapping, which is detailed in \cite{jiao2020robust}.
\textit{RHD02garden} is collected in a garden. Estimated trajectories and scene images are shown in Fig. \ref{fig.exp_rhd03traj}. 
Since the environment is well-conditioned, 
all trajectories are comparative.


\subsubsection{Localization Accuracy}
\label{sec.exp-per-accuracy}
We then perform a large-scale outdoor test on OR sequences under $10$ repetitions.
Environments on \textit{OR} sequences commonly provide sufficient features.
We visualize M-LOAM-gf's trajectory against the ground truth and the built map on \textit{OR02} in Fig. \ref{fig.exp_or02}, and plot trajectories the other sequences in Fig. \ref{fig.exp_or_others_trajectory}.
Each method is evaluated the absolute trajectory error (ATE) and the relative pose error (RPE) \cite{zhang2018tutorial}. 
Due to limited space, we report the translation ATE in Table \ref{tab.or_result}, and show complete results in our supplementary material \cite{jiao2020supplementary}.
M-LOAM-gf does not just preserve the accuracy of M-LOAM-full, but it also reduces the ATE on all sequences.
The average translation ATE of M-LOAM-gf is $22\%$ lower than that of  F-LOAM (the second-best method). 
The feature selection implicitly rejects outliers (see Section \ref{sec.gf-lslam}), which is essential to such accuracy gains.
Thus, M-LOAM-rnd also improves M-LOAM-full. But its drift is larger than that of M-LOAM-gf. 
The performance of the fully randomized operation in M-LOAM-rnd is not guaranteed, which occasionally lead to inconsistent results.

%


\subsubsection{Latency}
\label{sec.exp-per-or}

Experiments in the above sections are conducted on a desktop with an i7 CPU@4.2 GHz and 32 GB memory. 
The average latency of \textit{mapping} of M-LOAM-gf over $1725$ and $10356$ frames on the RHD and OR sequences is $62.69ms$ and $106.82ms$ respectively.
To demonstrate that our feature selection method boosts an L-SLAM  system on processors with limited resources, M-LOAM-gf is also tested on an Intel NUC\footnote{\url{zh.wikipedia.org/wiki/Next_Unit_of_Computing}} with an i7 CPU@3.1 GHz and 8 GB memory. 
The average latency is reported in Table \ref{tab.exp-latency-desktop}. 
We run the rosbag at a low frequency to ensure no data loss. 

First of all, we observe that M-LOAM-rnd has lower latency than M-LOAM-gf in the GF-based data association. This is because the rnd is an $O(M)$ algorithm, while the stochastic-greedy algorithm is $O(N\log(1/\epsilon))$.
Second, compared with M-LOAM-full, M-LOAM-gf may need more time for feature matching but significantly save time in nonlinear optimization.
Finally, M-LOAM-gf, M-LOAM-rnd, and LEGO-LOAM are three real-time systems ($<200ms$ at each mapping frame) for the Intel NUC.
Both M-LOAM-gf and M-LOAM-rnd outperform LEGO-LOAM in terms of accuracy.
LEGO-LOAM implicitly performs feature selection since it filters out points if the distances to their correspondences are larger than a threshold. But this naive and hard-coded solution leads to large accuracy loss.

\begin{table}[]
		\centering
		\renewcommand\arraystretch{1.25}
		\renewcommand\tabcolsep{3pt}  
		\caption{Average latency [ms] of mapping on an Intel NUC.}
		\begin{tabular}{|c|c|ccc|}
			\hline
			\multirow{2}{*}{Seq.} 
			& \multirow{2}{*}{Method} 
			& \multicolumn{3}{c|}{Mapping} \\
			
			\cline{3-5}
			& 
			& Data association
			& Optimization
			& Total \\ \hline
			
			\multirow{4}{*}{\textit{RHD}}
			& M-LOAM-gf   & $17.90$ & $3.09$ & $1\bm{15.48}$ \\
			& M-LOAM-rnd  & $6.33$ & $3.65$ & $\bm{92.57}$ \\
			& M-LOAM-full & $16.17$ & $7.01$ & $128.32$ \\
			& A-LOAM      & $-$ & $-$ & $131.08$ \\
			\cline{1-5}
			
			\multirow{6}{*}{\textit{OR}}
			& M-LOAM-gf   & $46.18$ & $4.64$ & $\bm{149.25}$ \\
			& M-LOAM-rnd  & $21.70$ & $6.86$ & $\bm{108.10}$ \\
			& M-LOAM-full & $54.35$ & $22.85$ & $230.35$ \\
			& A-LOAM      & $-$ & $-$ & $313.69$ \\
			& F-LOAM      & $-$ & $-$ & $271.30$ \\
			& LEGO-LOAM   & $-$ & $-$ & $158.23$ \\
			\hline
		\end{tabular}	
		\begin{tabular}{l}
			Latency: Time delay between the input and output of a function.
		\end{tabular}	
	\label{tab.exp-latency-desktop}		
	\vspace{-0.4cm}
\end{table}

\section{Conclusion}
\label{sec:conclusion}

In this paper, we propose a greedy-based feature selection method for NLS pose estimation using LiDARs. 
The feature selector retains the most valuable LiDAR features with the objective of preserving the information matrix's spectrum.
The stochastic-greedy algorithm is applied for the real-time selection.
Moreover, we also investigate the degeneracy issue of utilizing good features for pose estimation in structureless environments. 
We propose a strategy to adaptively change the number of good features to avoid ill-conditioned estimation.
The feature selection is integrated into a multi-LiDAR SLAM system, followed by evaluation on sequences with two sensor setups and computation platforms.
The enhanced system is shown to have great efficiency and higher localization accuracy than SOTA methods.
The idea of feature selection is general and can be applied to many NLS problems.

Future work will concern two possible directions.
The first direction is to utilize data-driven methods \cite{wong2020data} to
online tune parameters which were mannually set.
Another direction is to apply the proposed feature selection to other tasks, 
such as bundle adjustment \cite{zhao2020goodgraph} and cross-model localization \cite{huang2020gmmloc}.


\clearpage
\balance
\bibliographystyle{IEEEtran}
\bibliography{reference}{}

\end{document}